%% file: submission.tex
\documentclass{article} 
\usepackage{iclr2026_conference,times}

\input{math_commands.tex}

\usepackage{hyperref}
\usepackage{url}

\usepackage{microtype}
\usepackage{graphicx}
\usepackage{amssymb}
\usepackage{hyperref}
\usepackage{url}
\usepackage[capitalize]{cleveref}
\usepackage{multirow,makecell}
\usepackage{booktabs}
\definecolor{brickred}{RGB}{237,1,125}
\definecolor{olive}{RGB}{60,128,49}
\usepackage{xcolor}
\usepackage{subcaption}
\usepackage{listings}
\usepackage{wrapfig}

\usepackage{todonotes}

\crefname{section}{Sec.}{Secs.}
\Crefname{section}{Section}{Sections}
\Crefname{table}{Table}{Tables}
\crefname{table}{Tab.}{Tabs.}
\crefname{appendix}{Appx.}{Appxs.}
\crefname{figure}{Fig.}{Figs.}
\crefname{theorem}{Theorem}{Theorem}
\Crefname{lstlisting}{Listing}{Listings}
\definecolor{promptbackground}{gray}{0.95}
\lstdefinestyle{mystyle}{
    backgroundcolor=\color{promptbackground},
    commentstyle=\color{green},
    keywordstyle=\color{magenta},
    numberstyle=\tiny\color{gray},
    stringstyle=\color{purple},
    basicstyle=\ttfamily\footnotesize,
    breakatwhitespace=false,
    breaklines=true,
    captionpos=b,
    keepspaces=true,
    numbers=left,
    numbersep=5pt,
    showspaces=false,
    showstringspaces=false,
    showtabs=false,
    tabsize=2,
    literate={
      {√}{{$\scriptstyle^{\sqrt{}}$}}{1}
      {×}{{$\times$}}{1}
      {÷}{{$\div$}}{1}
    }
}
\usepackage{tcolorbox}
\tcbset{colback=gray!5!white, colframe=gray!50!black, boxrule=0.5pt, arc=2pt, before skip=8pt, after skip=8pt, top=1pt, bottom=1pt}

\newtcolorbox{promptbox}{colback=blue!5!white}
\newtcolorbox{outputbox}{colback=yellow!20!white}
\newtcolorbox{context}{colback=red!5!white}

\title{Focused Chain-of-Thought: Efficient LLM Reasoning via Structured Input Information}

\author{
Lukas Struppek$^{1 *}$
 \And Dominik Hintersdorf$^{2,3}$ \And Hannah Struppek$^{4}$ \AND \and \textbf{Daniel Neider}$^{5,6}$ \And Kristian Kersting$^{2,3,7,8}$\and \AND\\[0.0cm]
\vspace{-1cm}\\
$^{1}$FAR.AI, \\
$^{2}$German Research Center for Artificial Intelligence (DFKI), \\
$^{3}$Technical University of Darmstadt, \\
$^{4}$University of Kassel,\\
$^{5}$TU Dortmund University, \\
$^{6}$TU Center for Trustworthy Data Science and Security, University Alliance Ruhr, \\
$^{7}$Hessian Center for AI (Hessian.AI), \\
$^{8}$Centre for Cognitive Science, Technical University of Darmstadt\\
}

\iclrfinalcopy 
\begin{document}

\maketitle

\begin{abstract}
Recent large language models achieve strong reasoning performance by generating detailed chain-of-thought traces, but this often leads to excessive token use and high inference latency. Existing efficiency approaches typically focus on model-centric interventions, such as reinforcement learning or supervised fine-tuning, to reduce verbosity. In contrast, we propose a training-free, input-centric approach. Inspired by cognitive psychology, we introduce Focused Chain-of-Thought (F-CoT), which separates information extraction from the reasoning process. F-CoT first organizes the essential information from a query into a concise, structured context and then guides the model to reason exclusively over this context. By preventing attention to irrelevant details, F-CoT naturally produces shorter reasoning paths. On arithmetic word problems, F-CoT reduces generated tokens by 2–3× while maintaining accuracy comparable to standard zero-shot CoT. These results highlight structured input as a simple yet effective lever for more efficient LLM reasoning.
\end{abstract}

\input{sections/01_introduction}
\input{sections/02_related_work}
\input{sections/03_methodology}

\input{sections/04_experiments}
\input{sections/05_discussion}
\input{sections/06_conclusion}

\subsubsection*{Acknowledgments}
This work has been financially supported by the German Research Center for Artificial Intelligence (DFKI) project “SAINT”. We also gratefully acknowledge support from the hessian.AI Service Center (funded by the Federal Ministry of Research, Technology and Space, BMFTR, grant no. 16IS22091) and the hessian.AI Innovation Lab (funded by the Hessian Ministry for Digital Strategy and Innovation, grant no. S-DIW04/0013/003).

\bibliography{references}
\bibliographystyle{iclr2026_conference}

\appendix
\input{sections/A_appendix}
\input{sections/B_qualitative_examples}
\input{sections/C_quantitative_results}
\end{document}

%% file: math_commands.tex
\usepackage{amsmath,amsfonts,bm}

\def\eqref#1{equation~\ref{#1}}

\def\1{\bm{1}}

\DeclareMathAlphabet{\mathsfit}{\encodingdefault}{\sfdefault}{m}{sl}
\SetMathAlphabet{\mathsfit}{bold}{\encodingdefault}{\sfdefault}{bx}{n}



%% file: sections/01_introduction.tex
\section{Introduction}

Large language models (LLMs) are trained to predict the next token given a sequence of previous ones. Scaling model parameters and training data has substantially improved their performance on mathematical reasoning benchmarks, with recent models continuing to push the state of the art. Many LLMs reveal their internal reasoning by producing an explicit chain-of-thought (CoT)~\citep{wei2022chain} -- a step-by-step, natural-language rationale that makes reasoning traceable to humans. While CoT outputs increase transparency, they also generate long reasoning traces that are costly in time and computation. Moreover, locating errors within a long CoT is challenging, since the entire trace must typically be checked to identify the mistakes and verify correctness.
\renewcommand{\thefootnote}{}\footnote{$^*$Work mainly done at DFKI/Technical University of Darmstadt. Contact: \textit{FirstName}@far.ai.}\addtocounter{footnote}{-1}

The reasoning processes of LLMs are often compared to human logical thinking. Foundational work in cognitive psychology, such as the Active Control of Thought (ACT) framework~\citep{anderson1976language} models human problem-solving as sequential, resource-efficient processes, beginning with the representation and structuring of information before higher-order reasoning. Modern LLMs exhibit analogous reasoning behavior. However, the stages of information extraction and structuring in LLMs are not clearly distinguished from the subsequent reasoning phase and are often interwoven with it. We hypothesize that this entanglement blurs the boundaries between relevant and irrelevant information, thereby complicating the LLM reasoning process and contributing to the generation of unnecessary tokens. 

\begin{figure*}[t]
    \centering
    \includegraphics[width=\linewidth]{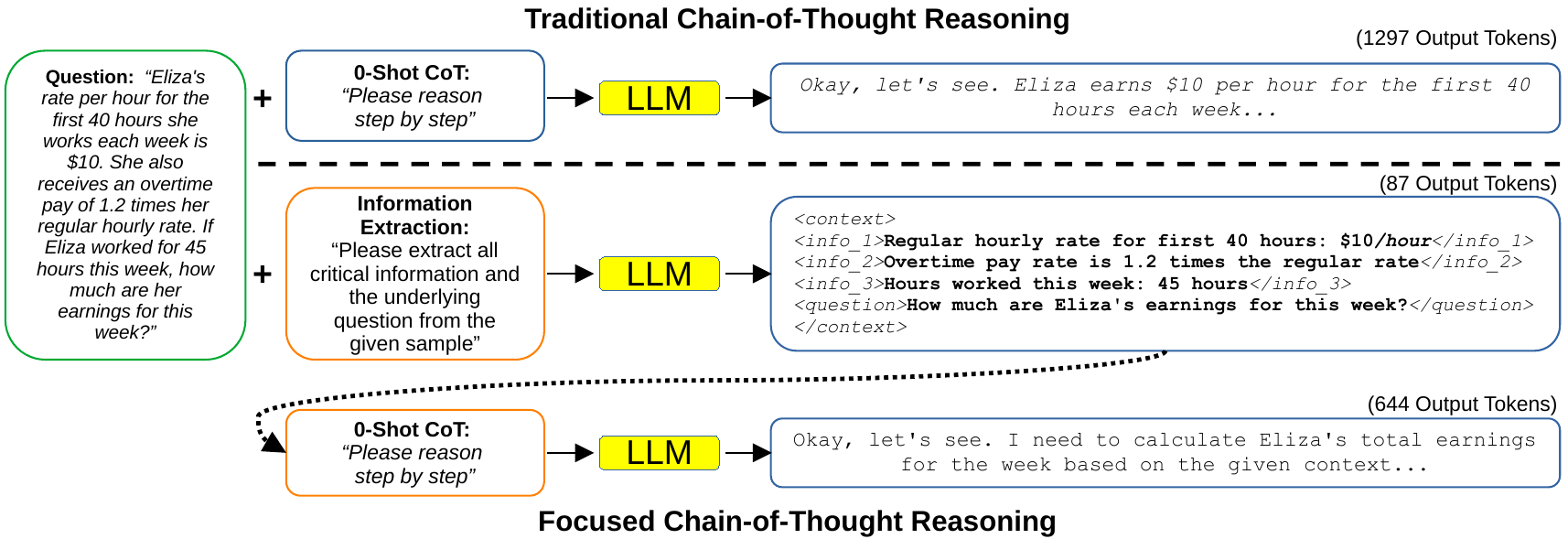}
    \caption{Focused Chain-of-Thought reasoning. The model first extracts key information into an XML-like context block and then performs reasoning based on that block. The context can also be pre-defined by the user or generated automatically by a larger LLM. When queried using only the context, large reasoning models produce significantly shorter reasoning traces compared to standard natural-language inputs. In this particular example, Qwen3 14B produces 43\% fewer tokens compared to standard CoT prompting. Shown prompts are abbreviated; see Appx.~\ref{appx:context_extraction} and \ref{appx:context_reasoning} for full prompts. }
    \label{fig:concept_overview}
\end{figure*}

In this paper, we introduce \textit{Focused Chain-of-Thought (F-CoT)}, a novel prompting strategy for reasoning tasks. Drawing inspiration from the ACT framework in cognitive psychology, F-CoT explicitly separates information extraction and structuring from the core reasoning process. In the first stage, we prompt an LLM to extract and organize relevant information from a question into a fixed, compact, and structured format. In the second stage, the model receives only this structured representation and performs standard chain-of-thought reasoning. A high-level overview of F-CoT is provided in \cref{fig:concept_overview}. By decoupling these two phases, F-CoT substantially reduces the number of generated tokens, accelerating inference by 2–3x compared to standard chain-of-thought prompting, while preserving strong reasoning performance. Crucially, our prompting strategy does not involve instructing or fine-tuning the LLM to shorten its reasoning---providing input information in a structured way is already sufficient to speed up reasoning. This makes F-CoT orthogonal to methods that explicitly encourage token efficiency through prompting strategies~\citep{xu2025chain,lee2025well}, supervised fine-tuning~\citep{yu2025long,luo2025autol2s}, or reward optimization~\citep{aggarwal2025l1, yeo2025demystifying}.

%% file: sections/02_related_work.tex
\section{Reasoning in Large Language Models}
LLMs are autoregressive models that generate text token by token. While recent models~\citep{dubey2024llama,deepseekai2025deepseekr1incentivizingreasoningcapability} have achieved impressive performance across diverse domains, unlocking their full potential requires more than large parameter counts and extensive training data. Their capabilities crucially depend on advanced inference strategies that elicit structured reasoning, i.e., the generation of logically consistent and interpretable intermediate steps rather than shallow text completions. This process is often compared to the slow, deliberative, and analytical (\textit{System 2}) mode of human thinking described by \citet{kahneman2011thinking}.

\textbf{Prompt-Based Reasoning.} A common way to trigger such reasoning behavior is to prompt the model to decompose complex problems into a series of intermediate steps, known as chain-of-thought (CoT) prompting~\citep{wei2022chain}. Notably, LLMs have been shown to reason effectively even in zero-shot settings, generating coherent reasoning steps without any in-context examples. Simply instructing a model to \textit{“think step by step”} can substantially enhance its reasoning performance~\citep{kojima2022large}. Building on these insights, subsequent work has proposed multi-path reasoning approaches such as tree-of-thought reasoning~\citep{yao2023tree}, self-consistency~\citep{wang2022self}, and ReAct~\citep{yao2023react}, which explore multiple reasoning paths before selecting or combining outcomes.

\textbf{Dedicated Reasoning LLMs.} An emerging class of specialized reasoning models, including the DeepSeek-R1~\citep{deepseekai2025deepseekr1incentivizingreasoningcapability} and Qwen3~\citep{qwen3technicalreport} families, employ a structured two-stage generation process. The model first enters a reasoning stage, often triggered by a special \textless think\textgreater\, token, followed by a concise summary stage where it synthesizes the final answer. These large reasoning models achieve state-of-the-art performance on logic and math benchmarks but often produce extremely long reasoning traces, sometimes spanning thousands of tokens for relatively simple questions~\citep{chen2024not,cuadron2025danger}. Even though these reasoning models perform better than traditional non-reasoning LLMs, their high parameter counts, combined with these lengthy reasoning processes significantly increase inference time and computational cost.

\textbf{Test-Time Scaling.} However, this relationship of improved performance due to longer reasoning traces can also be exploited by allocating more inference time, typically by generating additional tokens, commonly referred to as test-time scaling. Various inference-time techniques implement this principle by extending or parallelizing the reasoning process, for example, by using reconsideration tokens~\citep{muennighoff2025s1} or self-refinement steps~\citep{madaan2023self,tian2025think}. While such approaches aim to enhance accuracy by scaling up inference compute, the opposite direction, i.e., reducing the number of generated tokens to save time and resources, has received comparatively less attention, even though large reasoning models are known to suffer from overthinking~\citep{chen2024not,chiang2024over}.

\textbf{Efficient Reasoning.} Existing approaches for reducing token count during reasoning typically intervene at the training or fine-tuning stage. Reinforcement learning methods explicitly reward shorter outputs~\citep{aggarwal2025l1,yeo2025demystifying,arora2025training,ye2025emergence}. Data-centric approaches, on the other hand, shorten reasoning traces either by removing redundant and irrelevant tokens~\citep{zhuang2025accelerating,xia2025tokenskip,yuan2025not,xiao2025limopro} or by fine-tuning on shorter chain-of-thought examples~\citep{yu2025long,luo2025autol2s}. Existing training-free approaches aim to reduce the token count by explicitly instructing the model to keep its reasoning short and concise~\citep{xu2025chain,nayab2024concise}.

All these methods for increasing reasoning efficiency have in common that they target the model itself, either by explicitly instructing or fine-tuning the model to produce shorter reasoning traces. In contrast, we adopt a fundamentally different perspective that targets the input rather than the model. With our Focused Chain-of-Thought (F-CoT) framework, we demonstrate that presenting information in a structured and compact form naturally leads to shorter reasoning traces and substantial token savings, without any additional instructions or fine-tuning. 
While this approach is related to chain-of-thought prompting~\citep{wei2022chain}, as it still encourages sequential reasoning, it fundamentally differs in how the input information is represented. Similarly, it differentiates itself from schema-based prompting~\citep{zhong2023improving}, which organize outputs in structured formats such as JSON or XML, as F-CoT instead structures inputs to enhance reasoning efficiency. Moreover, while F-CoT has similarities with retrieval-augmented generation~\citep{lewis2020retrieval} as it separates information retrieval from reasoning, it performs self-extraction instead of querying an external database. Notably, while knowledge-graph approaches~\citep{pan2024unifying,kau2024combining,ma2025large} also leverage structured factors, F-CoT fundamentally differs by relying solely on self-contained, compact representations for reasoning, rather than externally provided knowledge.

%% file: sections/03_methodology.tex
\section{From Natural Language to Structured Representations}
Natural language descriptions in mathematical reasoning tasks often contain the essential facts alongside irrelevant wording. This unstructured form of information can lead LLMs to produce long, verbose reasoning traces. Consider the following arithmetic problem:
\begin{promptbox}
\textbf{Prompt:} An apple costs \$2 in the supermarket. A pear costs one and a half times the price of an apple. Eve wants to buy some fruits for herself to eat more healthily. What is the total cost of two apples and two pears?
\end{promptbox}

When given such a question, an LLM typically begins by interpreting the text and directly reasoning about it:

\begin{outputbox}
\textbf{LLM Output:} Okay, let me try to figure out this problem. The question is about finding the total cost of two apples and two pears in the supermarket. [...] First, it says an apple costs \$2. [...] Then, a pear costs one and a half times the price of an apple. [...]
\end{outputbox}

This natural-language-based reasoning process generates many unnecessary tokens and scatters key information across verbose text, even when the facts are simple. By contrast, humans are taught in school first to extract and note down the essential facts in a structured format. For example, giving the same problem to a person, one would start by identifying all the necessary information for solving the task:
\begin{context}
\textbf{Info 1:} Price per apple: \$2\\
\textbf{Info 2:} Price per pear: $1.5 \times$ price per apple
\end{context}
A structured representation allows humans to attend to key information more easily, since it is no longer buried in full sentences and irrelevant information but isolated in a clear and concise format. This compact representation isolates the relevant details, making reasoning easier and reducing cognitive overhead. Inspired by this human strategy, we hypothesize that providing LLMs with similarly structured inputs can shorten reasoning paths, reduce irrelevant attention distraction, and improve efficiency and faithfulness.

\subsection{From Text to Structured Facts}
To test this hypothesis, we define a fixed, structured context format that explicitly separates factual information from the question:

\begin{context}
\textless \textbf{context}\textgreater\\
    $\phantom{0000}$\textless \textbf{info\_1}\textgreater Price per apple: \$2\textless \textbf{/info\_1}\textgreater \\
    $\phantom{0000}$\textless \textbf{info\_2}\textgreater Price per pear: 1.5x price per apple\textless \textbf{/info\_2}\textgreater \\
    $\phantom{0000}$\textless \textbf{question}\textgreater Total cost of 2 apples and 2 pears\textless \textbf{/question}\textgreater\\
\textless /\textbf{context}\textgreater
\end{context}

This structure contains two components: (1) enumerated information blocks \textless info\_k\textgreater...\textless /info\_k\textgreater, and (2) a \textless question\textgreater\;field specifying the objective. The XML-like format supports automatic parsing, syntax validation, and compatibility with downstream pipelines.

While we use this XML-like format for clarity, our experiments (see \cref{sec:ablation}) show that simpler enumerated lists yield comparable performance, suggesting that the key benefit arises from structure itself, not from the specific syntax.

\subsection{Implementing Structured Reasoning Pipelines}
There are two practical strategies to supply structured information in our F-CoT framework:
\begin{enumerate}
\item \textbf{Pre-formatting by the user:} The user manually provides the question in the defined structure, allowing the model to focus entirely on reasoning. This minimizes model workload but requires additional user effort.
\item \textbf{Two-step prompting:} The model first extracts the context (without reasoning) and then reasons over it. For less capable LLMs, extraction may be unreliable; in such cases, a larger model can generate the context, which a smaller model then uses for reasoning. This hybrid pipeline combines the extraction capabilities of large models with the efficiency of smaller ones, reducing total inference cost. The two-step prompting approach is also depicted in \cref{fig:concept_overview}.
\end{enumerate}

Throughout this paper, we follow the second approach, which first queries an LLM to extract and generate the context for a given question. We then provide only this extracted context, without the original question, to the model, asking it to reason about it. Our reasoning prompt extends zero-shot chain-of-thought prompting by explicitly instructing the model to (i) focus solely on the structured context, (ii) reason step by step, and (iii) cite relevant \textless info\_k\textgreater\ blocks when using them. These explicit citations make the reasoning process interpretable and facilitate debugging. We present the exact prompts used for F-CoT in \cref{appx:context_extraction}.

%% file: sections/04_experiments.tex
\section{Experiments}
In the following, we experimentally demonstrate that our proposed F-CoT matches the accuracy of traditional zero-shot CoT reasoning~\citep{kojima2022large} with state-of-the-art reasoning LLMs on standard math benchmarks, while producing far more concise reasoning.

\subsection{Experimental Protocol}
\textbf{LLMs and Hyperparameters:} 
We use 
\textit{Qwen-3} (0.6B, 4B, 14B, 32B)~\citep{qwen3technicalreport} reasoning LLMs, i.e., models with an explicit thinking mode. For all models, we use sampling parameters recommended by the model developers: temperature 0.6, top-p 0.95, min-p 0.0, and top-k 20. The maximum number of generated tokens is set to 32k, matching the context length supported by Qwen-3 models. When using the LLMs to generate the context, we skip the reasoning process for Qwen-3 models by appending a \textit{no\_think} flag to the input prompt. 

\textbf{Evaluation Benchmarks:} We focus on arithmetic word problems, i.e., problems that combine key information with unrelated details such as descriptions of settings, names, and situations. We primarily rely on three datasets with varying levels of difficulty: \textit{SVAMP}, \textit{GSM-Hard}, and \textit{MATH-500}. The SVAMP dataset~\citep{patel2021nlp} is comparably easy, containing short and straightforward questions. GSM-Hard is a more challenging version of the GSM8k~\citep{cobbe2021gsm8k} test set of grade-school math problems. The hard version~\citep{gao2023pal} replaces the original numbers with less common ones, thereby increasing the difficulty of the reasoning process and reducing the risk of test set contamination. In addition, we employ the more demanding MATH-500 benchmark~\citep{hendrycksmath2021}. While our main focus is on verbose mathematical questions, we also analyze the impact of highly condensed problems using the AIME2024 and AIME2025 datasets~\citep{aime2025}. Since these problems are not the central focus of this paper, we report the corresponding results in the appendix and discuss only the key findings in the main text.

\textbf{Metrics:} We compute the Pass@5 metric, which evaluates whether at least one of the five outputs generated by a model for a given problem is correct. Additionally, we calculate the average number of tokens generated per question, denoted as \textit{\# Tokens}. For settings using our F-CoT reasoning where the LLM generates the context itself, this token count also includes tokens from the extracted context to ensure a fair comparison. For self-generated contexts, we further measure the proportion of valid context blocks generated, i.e., whether the context adheres to a valid XML structure.

\textbf{Baselines:} We compare the F-CoT prompting strategy against the standard zero-shot CoT prompting without any contextual information provided (\textit{0-CoT}). As additional baselines, we investigate Plan-and-Solve prompting (\textit{P\&S})~\citep{wang2023plan} and Condition-Retrieving Instruction prompting (\textit{CoRe})~\citep{coleg_xu}, two prompting strategies designed to improve 0-CoT prompting by asking the model to first identify crucial conditions and objectives before starting the reasoning process. These methods follow a strategy similar to our F-CoT, but without explicitly separating the two steps or enforcing a fixed structure. Since we do not explicitly instruct the model to reduce its reasoning effort, e.g., by employing reinforcement learning-based fine-tuning to produce fewer tokens, we do not compare against such approaches. Instead, our goal is to investigate how the availability of structured information benefits the reasoning process.

\begin{figure*}[t]
    \centering
    \includegraphics[width=\linewidth]{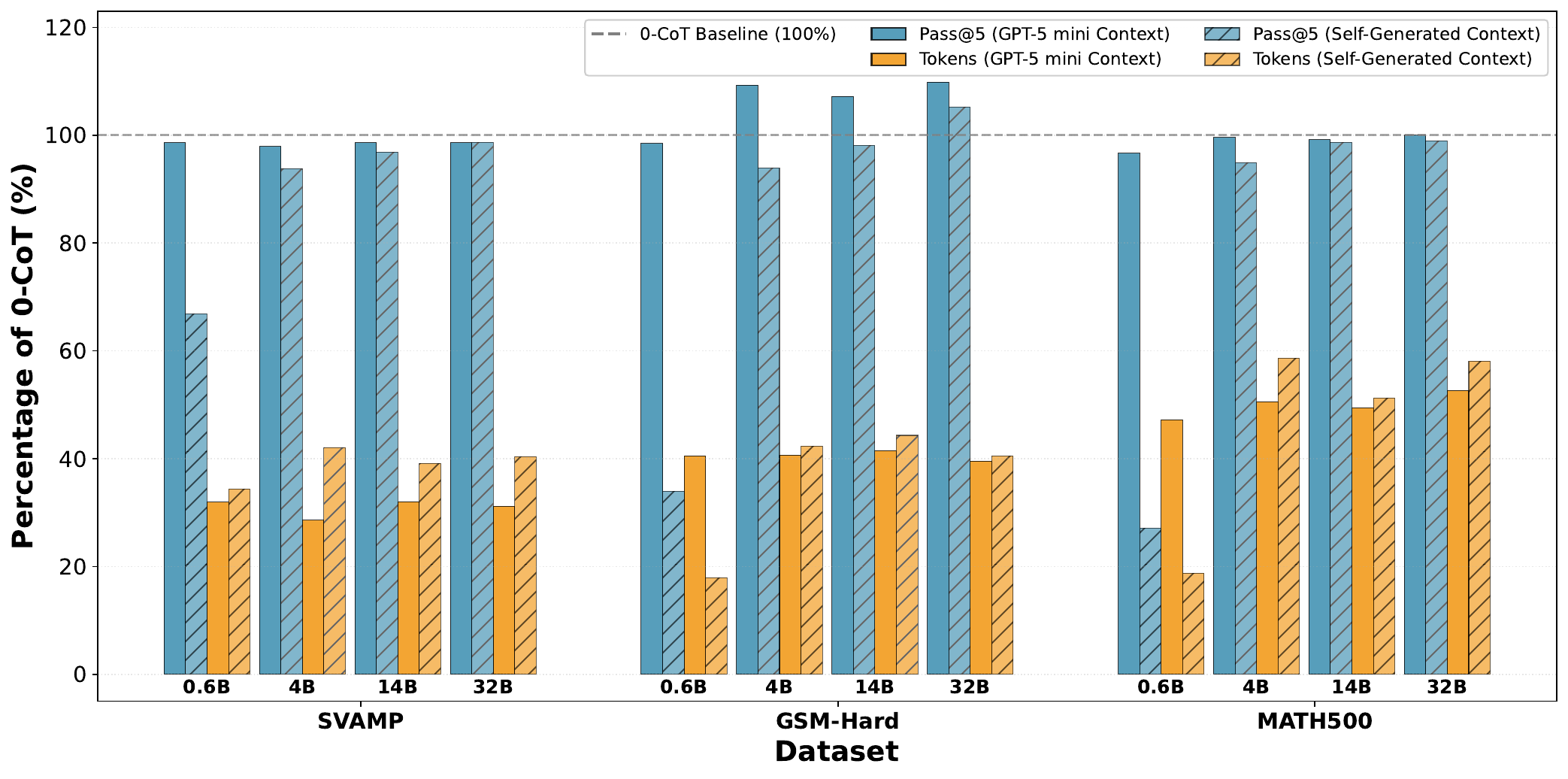}
    \caption{Comparison of 0-CoT and our F-CoT using Qwen3 models of various sizes. For F-CoT, two settings are shown: context pre-computed by GPT-5 mini (solid bars) and generated by the model itself (hatched bars). F-CoT results are expressed relative to 0-CoT. While F-CoT matches 0-CoT performance in most cases, it generates substantially fewer tokens, thereby improving inference efficiency. Detailed numerical results are provided in Appendices~\ref{appx:precomputed_context_results} and \ref{appx:self_generated_context_results}.}
    \label{fig:precomputed_context}
\end{figure*}

\subsection{Boosting Reasoning Efficiency with Pre-Computed Context}
We begin by considering the setting in which the user provides the LLM with a pre-computed context. While transforming a question into such a context can be done manually, we use \textit{GPT-5 mini} (version: \textit{2025-08-07}), a state-of-the-art LLM, for this task. These experiments aim to evaluate the impact of structured information on the reasoning process of LLMs. The LLMs’ ability to generate context autonomously is assessed in the next subsection.

For context extraction, we instructed GPT-5 to include only information directly relevant to answering the question, while ignoring background details. The model was explicitly asked to refrain from solving the problem or performing calculations, to avoid evaluation biases stemming from pre-computed reasoning steps. Additionally, we provided in-context two examples of correctly extracted contexts to ensure proper formatting. Full prompts are available in \cref{appx:gpt5_prompts}. The prompts for GSM-Hard/SVAMP and MATH differ only in the examples provided, reflecting the differences in question style. On average, GPT-5 generated contexts for MATH-500 consist of 126 tokens, meaning context generation introduces minimal overhead.

The evaluation results are shown in \cref{fig:precomputed_context} (non-hatched bars). We report the F-CoT results as relative values compared to zero-shot CoT (0-CoT). For example, a token usage of 40\% indicates that the model required only 40\% of the tokens on average compared to 0-CoT. Absolute metrics are additionally provided in \cref{tab:results_precomputed} in \cref{appx:precomputed_context_results}.

The most notable finding is that providing reasoning LLMs with context substantially reduces the number of generated tokens (orange bars). For SVAMP questions, token counts drop to roughly one-third of 0-shot CoT across all model sizes. For more challenging GSM-Hard and MATH-500 questions, token counts decrease by about half. For example, Qwen3 32B generates only around 2.3k tokens on MATH-500 questions when provided with the context, compared to roughly 4.4k tokens under 0-shot CoT, while achieving the same Pass@5 scores. This reduction translates to a significant inference speed-up, and we observe similar token savings across all model sizes.

In terms of reasoning accuracy (blue bars), performing reasoning over the provided context yields results comparable to 0-CoT across all three datasets, except for the smallest Qwen3 0.6B model, which exhibits slightly lower performance. Larger models, in contrast, outperform 0-shot CoT on GSM-Hard. When manually analyzing the results, we find that the main differences arise in questions requiring negative results. With 0-CoT, models sometimes misinterpret large numbers due to hallucinated/confabulated typos, producing incorrect answers. When reasoning over a structured context, the models still note unusually large numbers but avoid false interpretations, adhering to the provided information. An example reasoning trace is shown in \cref{appx:gsm_hard_comparison}. However, since careful prompting could mitigate such errors, 0-CoT performance could likely match structured context results if explicitly instructed to interpret numbers as given.

Prompting the models with \textit{Plan-and-Solve} and \textit{CoRe} does not lead to a noticeable difference, neither in reasoning capabilities nor in the number of generated tokens -- both values are comparable to the results with 0-CoT prompting. We, therefore, report the corresponding results only in \cref{appx:precomputed_context_results}.

For AIME questions, which already present condensed information and do not require prior extraction, we still observe a substantial reduction in token usage when using structured context. However, correctness slightly decreases. Given the small dataset size (30 samples) and potential test-set contamination~\citep{golchin2023time,oren2023proving}, we cannot conclusively claim that F-CoT harms reasoning in this setting. The observed difference may instead reflect that the model has likely seen the original questions during training, whereas the structured context representation is novel, a phenomenon that also applies to other benchmarks.

\subsection{Context Generation and Reasoning Within a Single Model}
So far, we have explored F-CoT by providing the model with external information in the form of pre-computed context generated by another LLM (GPT-5 mini). We now focus on the capabilities of vanilla reasoning LLMs to generate such contexts themselves. In other words, following the concept in \cref{fig:concept_overview}, both steps, i.e., context generation and reasoning, are performed by the same model. As before, models see only the generated context during the reasoning step, with no access to the original input. The prompt used for context generation is provided in \cref{appx:context_extraction}, and the relative results are again reported in \cref{fig:precomputed_context} (hatched bars). See \cref{appx:self_generated_context_results} for numerical results. Unlike the context generation with GPT, where the prompt includes some examples of successfully generated contexts, we do not provide such few-shot demonstrations to offer the models more flexibility in their context generation and to obtain a clearer image of their inherent capabilities in generating contexts.

Regarding token counts (including context tokens), we observe values comparable to those in the previous section, again highlighting a substantial reduction compared to 0-shot CoT. In terms of reasoning performance, larger models tend to perform better, often matching the results achieved with GPT-5 pre-computed contexts. However, smaller models, particularly Qwen3 0.6B, exhibit a notable drop in performance. This decline is primarily due to their inability to reliably extract and structure information in the desired format: fewer than 2\% of generations produce a valid context. While some of these contexts still contain enough information to solve the question, the model can answer only a limited subset of problems, underperforming compared to both 0-shot CoT and externally computed contexts. Context generation is more stable for Qwen3 14B and larger, where almost all generations produce valid contexts.

These results suggest a practical and cost-efficient strategy for F-CoT: pre-compute the context, which typically requires only a few tokens, using a larger, slower, and more expensive LLM, and then perform the token-intensive reasoning step with a smaller, faster, and more efficient model.

\subsection{Quantifying and Characterizing Reasoning Dynamics}
Next, we examine how F-CoT impacts the reasoning dynamics of LLMs. We focus in this section on the Qwen3 14B model and its results on MATH-500 since the model offers a good balance between reasoning capabilities and size. Moreover, the MATH-500 benchmark offers a diverse range of questions.

We first adapt the overthinking score originally proposed by \citet{cuadron2025danger} to quantify overthinking in reasoning LLMs on agentic tasks. In our study, overthinking refers to a model’s tendency to prioritize abstract reflection, speculation, or multi-branch planning over performing the concrete mathematical steps that directly advance the solution. The overthinking score, ranging from 0 to 10, measures the degree to which a model emphasizes internal deliberation over concrete problem-solving. Low scores indicate focused, step-by-step reasoning with minimal speculative discussion, while high scores reflect extensive meta-reasoning, frequent method-switching, or premature conclusions without deriving intermediate results.

To compute the score, we modify the original prompt to emphasize mathematical reasoning rather than general agentic behavior. Sample-wise scores are then predicted by GPT-5 Mini, with the full prompt provided in \cref{appx:overthinking_score}. The overthinking score for the model using 0-shot CoT is $2.35 \pm 1.5$, indicating mild overthinking. Incorporating the context reduces the score to $1.74 \pm 1.4$, demonstrating improved reasoning efficiency. Moreover, the score decreases in $50.2\%$ of the samples and remains unchanged in $38.4\%$, highlighting consistent reductions in unnecessary reasoning.

\begin{figure*}[t]
    \centering
    \includegraphics[width=0.9\linewidth]{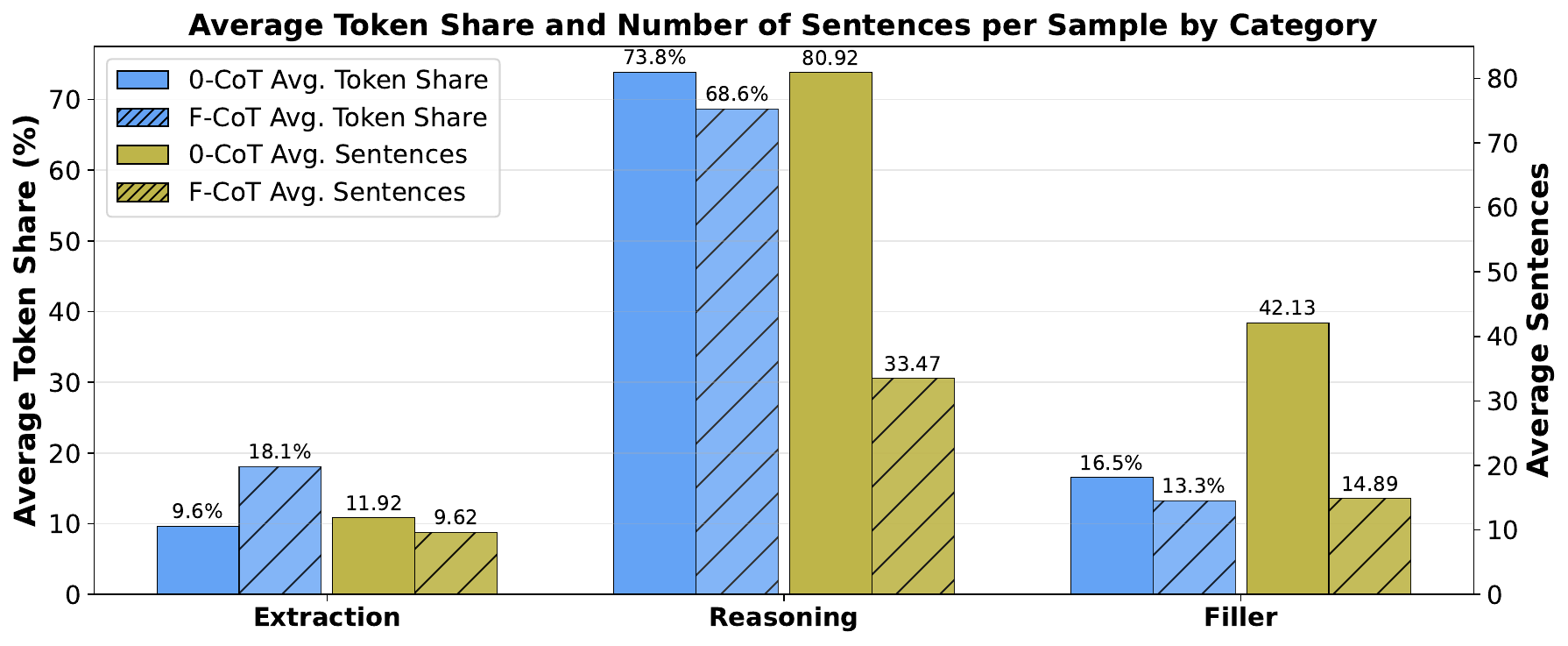}
    \caption{Analysis of reasoning traces during the chain-of-thought, where each sentence is classified as \textit{Extraction}, \textit{Reasoning}, or \textit{Filler}. Blue bars indicate the average share of tokens per category, while green bars show the average number of sentences per category. Although the relative token distribution remains largely unchanged, the number of reasoning and filler sentences is substantially reduced when using our F-CoT compared to 0-CoT. Model: Qwen3-14B; Dataset: MATH-500.}
    \label{fig:token_count}
\end{figure*}

We further annotate each output sentence of the Qwen3 model using GPT-5 Mini as \textit{Extraction}, \textit{Reasoning}, or \textit{Filler}. Extraction sentences copy or paraphrase information from the input question without transformation or inference. Reasoning sentences involve logical or arithmetic inference or provide explanations not explicitly stated in the input. Filler sentences include remaining utterances that do not contribute to the solution, such as phrases like \textit{“Wait, let me re-read the problem.”} or \textit{"So that makes sense."} We then compare the token distribution across the different categories, as well as the average number of sentences classified as each category.

\Cref{fig:token_count} visualizes the results. When comparing the token distribution across categories (blue bars), no fundamental difference is apparent between standard 0-CoT and our F-CoT. A slight token increase is observed for F-CoT in the extraction category. However, this can be attributed to the model explicitly referencing individual information blocks during reasoning, which are consequently labeled as extraction. In terms of the average number of sentences per category (green bars), both prompting strategies produce roughly the same number of extraction sentences. In contrast, F-CoT more than halves the number of reasoning sentences and substantially reduces the number of filler sentences, which do not directly contribute to solving the problem. We therefore conclude that the overall token reduction achieved by F-CoT primarily results from fewer reasoning and filler sentences, while the relative distribution of tokens across categories remains largely unchanged.

\subsection{Ablation Study and Sensitivity Analysis}\label{sec:ablation}
In the final experimental section, we want to clarify the impact of individual design choices for our F-CoT experiments. We again focus on Qwen3 14B and the MATH-500 dataset. We report exact numerical results in \cref{tab:sensitivity_results} in \cref{appx:sensitivity_results}

\textbf{Q1: Does the prompt format influence the token count?} 
Previous research~\citep{sclar2024quantifying,errica2025did} has pointed out the sensitivity of LLMs to their prompt design. To ensure that the structured information is the driving force behind the shorter reasoning traces, we repeat the experiments on MATH-500 using the Qwen3 14B model. We minimize the details in our LLM instruction to use the context to a minimum, followed by the standard 0-shot CoT prompt after the context. The shortened prompt is presented in \cref{lst:reasoning_generation_short} in \cref{appx:context_reasoning}. Removing the explicit instructions on how to use the context increases the average token count to about 3.1k (+747 tokens compared to default F-CoT) while also slightly increasing the pass@5 to 99.2\% (+0.6pp). 

We repeat this experiment using the same prompt with the standard questions instead of the compact context representation to see whether it also affects these results. Indeed, we observe a token count of 4.2k ($-$685 tokens compared to default 0-CoT) and also a pass@5 accuracy of 99.2\%. Yet, there remains a clear gap of more than 1k tokens between the inference with and without the context, even when using the same prompt type. We also report this prompt in \cref{lst:adjusted_0_cot_prompt} in \cref{appx:cot_reasoning}.

\textbf{Q2: Does the context format influence the results?}
We further assess whether the format of the context plays a crucial role. Instead of using the XML format, we explore three alternative design choices that abandon this strict structure. First, we use a simple \emph{enumerated list} of information items followed by the question. Second, we employ an \emph{unnumbered list} to present the information. Finally, we completely remove any structured representation and simply concatenate all information, followed by the question. The structures of all three context variations are illustrated in \cref{appx:context_designs}.

We observe no difference in Pass@5 or token count when provided with the context as an enumerated list. While performance remains unchanged for the unnumbered list, the token count increases by 12.6\%. A similar effect is observed when concatenating all information without structure. We therefore conclude that the exact format or representation of the extracted information is not crucial for the success of our F-CoT. All forms of structured and compressed information substantially reduce the number of generated tokens. However, introducing more structure, such as numbering individual information pieces, further decreases the token count.

\textbf{Q3: Can a model benefit from receiving both the original question and the extracted context?}
While we implicitly assume that the provided context contains all information necessary to answer a question, including the original question alongside the extracted context may offer additional benefits. To test this, we repeat the inference process by providing both the structured context and the original question. The model is instructed to focus primarily on the context and refer to the original question only if certain information appears unclear or ambiguous. The adjusted prompt is shown in \cref{lst:context_and_question} in \cref{appx:context_reasoning}.

When queried with both the context and the original question, the model achieves a slight improvement in Pass@5 (+0.8 percentage points), accompanied by a 19\% increase in token count. Nevertheless, the total number of generated tokens remains substantially lower than in the 0-CoT setting, approximately 41\% fewer tokens overall. We conclude that this prompting strategy can be beneficial in scenarios where precomputing a precise context is challenging due to unclear phrasing or potentially ambiguous interpretation of the question.

\textbf{Q4: Do larger models compute better contexts?}
We further examine whether larger models produce more reliable context information than smaller ones. To evaluate this, we generate context for all MATH-500 samples using different Qwen 3 model sizes (ranging from 0.6B to 32B), and then perform the F-CoT reasoning step with the Qwen 3 14B model.

As shown in \cref{tab:sensitivity_results_model_size} in \cref{appx:sensitivity_results}, larger models ($\geq$14B) indeed produce higher-quality context, leading to improved performance. However, these gains saturate at the 14B scale, and we do not expect further improvements beyond this point. The experiments also show that models with 4B parameters or more consistently generate valid XML-like context, whereas the 1.7B model does so in about 95\% of cases. Even though the 0.6B model almost always produces context that is invalid from a strict XML standpoint, the 14B model still achieves a Pass@5 accuracy of 41.2\%, indicating that perfect XML format validity is not strictly necessary for correct reasoning.

%% file: sections/05_discussion.tex
\section{Discussion}
We finish the paper with a general discussion of the limitations and possible future directions.

\subsection{Limitations}
When automatically creating context windows with LLMs, information loss may occasionally occur during the extraction process. For example, we observed rare cases in which instructions such as “state the answer in percentage values” or other details critical for interpreting the input are omitted, causing the model to output decimal numbers instead (e.g., 0.05 instead of 5\%). Such discrepancies can negatively impact performance on math benchmarks.

In other cases, the extracted information may be misinterpreted, which can subsequently affect the downstream reasoning process. Thanks to the structured format of our context, however, such issues can be easily identified and corrected. We also found that transforming already highly condensed information, such as AIME questions, into our context scheme does not improve reasoning correctness and can sometimes even degrade it. Moreover, smaller models, particularly the 0.6B variant, struggle with self-generated context extraction. In these cases, a larger model must be used to pre-compute the context to achieve speedups without significantly harming the smaller model’s capabilities.

\subsection{Future Directions}
Until now, we have explored F-CoT in isolation from other prompting techniques. However, combining a structured representation of input information with more advanced prompting strategies could further reduce the number of generated tokens or improve performance. Another promising avenue is to integrate F-CoT with test-time scaling methods such as tree-of-thoughts, potentially accelerating them while enabling richer exploration.

Although our focus is on reasoning-oriented language models, we believe that similar structured information extraction could also benefit multimodal models. For example, when querying a vision-language model with an image, first extracting key visual elements, either in a structured format or as image excerpts, before performing higher-level reasoning may yield efficiency gains similar to those observed in our setting. Another direction is to investigate how structured reasoning can be incorporated directly into model training or fine-tuning, improving the model’s inherent ability to identify and extract salient information. Currently, the evaluated models were not trained with such structured inputs, which may limit their native capacity to process them.

We also envision using the context as a dynamic notepad, where the model can store intermediate reasoning steps by updating existing entries or adding new ones. Thanks to the fixed XML structure, context updates can be introduced incrementally. Adjusting the model’s prefix to reflect these updates could help concentrate the model’s attention on the curated context rather than on thousands of previously generated tokens. However, this approach would likely require additional fine-tuning and would necessitate recomputing cached key–value states whenever the context is modified.

%% file: sections/06_conclusion.tex
\section{Conclusion}
We present Focused Chain-of-Thought (F-CoT), a simple yet highly effective prompting strategy that explicitly separates information extraction from the reasoning process. By querying reasoning LLMs with structured context rather than raw natural language, F-CoT addresses the issue of excessive token usage, achieving a 2-3x reduction in generated tokens while preserving reasoning performance. Our analysis reveals that this efficiency stems from a reduction in filler and redundant reasoning steps, as structured inputs naturally mitigate the model's tendency to overthink. Ultimately, these results demonstrate that optimizing input representations offers a powerful, training-free alternative to model-centric optimization for efficient deployment.

%% file: sections/A_appendix.tex
\newpage
\section{Prompt Designs}


\subsection{Context Extraction with LLMs without Examples}\label{appx:context_extraction}

In our experiment, we used the following prompt shown in Listing~\ref{lst:context_extraction} to instruct the model to extract context information from input data.

\begin{lstlisting}[
  caption={The prompt used to generate context information.},
  label={lst:context_extraction}
]
Please extract all critical information and the underlying question from the given sample.
The output must follow this format:
<context>
<info_1>Information 1</info_1>
<info_2>Information 2</info_2>
<question>The question that needs to be answered based on the information provided.</question>
</context>

Guidelines:
- Only include information directly relevant to answering the question.
- Write information in concise, factual statements.
- If quantities, measurements, or mathematical relations are given, include them (use LaTeX for numbers and units when appropriate, e.g., $4.5 \\, \\text{inches}$).
- Ignore irrelevant background or narrative details.
- Restate the question clearly and concisely, keeping only what is necessary to answer it.
- Do not solve the problem or perform calculations.

Please extract the information and question from the following sample:

[ORIGINAL QUESTION]
\end{lstlisting}

\newpage 
\subsection{Context Generation with GPT-5-mini}\label{appx:gpt5_prompts}

In our experiment, we used the prompt shown in Listing~\ref{lst:prompt_gsm_extraction} to instruct the model to extract context information from the GSM-hard and SVAMP data. Analogously, the prompt in Listing~\ref{lst:prompt_math_extraction} is used to extract context information from MATH-500 and AIME data.

\begin{lstlisting}[
  caption={The prompt used to generate context information from GSM-hard and SVAMP samples. Model: GPT-5 Mini (August 2025 version).},
  label={lst:prompt_gsm_extraction}
]
Please extract all critical information and the underlying question from the given sample.
The output must follow this format:
<context>
<info_1>Information 1</info_1>
<info_2>Information 2</info_2>
<question>The question that needs to be answered based on the information provided.</question>
</context>

Guidelines:
- Only include information directly relevant to answering the question.
- Write information in concise, factual statements
- If quantities, measurements, or mathematical relations are given, include them (use LaTeX for numbers and units when appropriate, e.g., $4.5 , \text{inches}$).
- Ignore background or story-like details.
- Restate the question in a minimal, clear form. Avoid repeating irrelevant phrasing.
- Do not solve the problem or perform calculations.

Here is an example:
Question: Three of the women at the cocktail party are wearing 4.5-inch heels and three are wearing 2.5-inch heels. What is the average height of heels at this party?

Desired Output:
<context>
<info_1>Three women wearing 4.5-inch heels</info_1>
<info_2>Three women wearing 2.5-inch heels</info_2>
<question>What is the average heel height?</question>
</context>

Here is another example:
Question: Fishio posted her selfie on Instagram. She received 20000 likes on the photo after 1 week. Three weeks later, the number of likes was 70 times as many as the initial number of likes. If she received 200000 more new likes recently, how many Instagram likes are there?

Desired Output:
<context>
<info_1>Initial likes after 1 week: 20000</info_1>
<info_2>Likes after 3 weeks: \(70(20000)\)</info_2>
<info_3>Additional $200000$ likes received after the 3-week count</info_3>
<question>How many Instagram likes are there?</question>
</context>

Please extract the information and question from the following sample: 

[ORIGINAL QUESTION]
\end{lstlisting}

\newpage 

\begin{lstlisting}[
  caption={The prompt used to generate context information from MATH-500 and AIME samples. Model: GPT-5 Mini (August 2025 version).},
  label={lst:prompt_math_extraction}
]

Please extract all critical information and the underlying question from the given sample.

The output must follow this format:

<context>

<info_1>Information 1</info_1>

<info_2>Information 2</info_2>

<question>The question that needs to be answered based on the information provided.</question>

</context>

Guidelines:

- Only include information directly relevant to answering the question.

- Write information in concise, factual statements

- If quantities, measurements, or mathematical relations are given, include them (use LaTeX for numbers and units when appropriate, e.g., $4.5 , \text{inches}$).

- Ignore background or story-like details.

- Restate the question in a minimal, clear form. Avoid repeating irrelevant phrasing.

- Do not solve the problem or perform calculations.

Here is an example:

Question: Let \\[f(x) = \\left\\{\n\\begin{array}{cl} ax+3, &\\text{ if }x>2, \\\\\nx-5 &\\text{ if } -2 \\le x \\le 2, \\\\\n2x-b &\\text{ if } x <-2.\n\\end{array}\n\\right.\\]Find $a+b$ if the piecewise function is continuous (which means that its graph can be drawn without lifting your pencil from the paper).

Desired Output:

<context>

<info_1>Piecewise function: 

\[

f(x) = 

\\begin{cases} 

ax + 3, & 	ext{if } x > 2 \

x - 5, & 	ext{if } -2 \le x \le 2 \

2x - b, & 	ext{if } x < -2

\end{cases}

\]</info_1>

<info_2>The function is continuous (can be drawn without lifting the pencil)</info_2>

<question>Find $a+b$.</question>

</context>

Here is another example:

Question: A rectangular band formation is a formation with $m$ band members in each of $r$ rows, where $m$ and $r$ are integers. A particular band has less than 100 band members. The director arranges them in a rectangular formation and finds that he has two members left over. If he increases the number of members in each row by 1 and reduces the number of rows by 2, there are exactly enough places in the new formation for each band member. What is the largest number of members the band could have?

Desired Output:

<context>

<info_1>Band has less than 100 members</info_1>

<info_2>Original formation: $m$ members per row, $r$ rows, with 2 members left over</info_2>

<info_3>New formation: $(m+1)$ members per row, $(r-2)$ rows, exactly enough places for all members</info_3>

<question>What is the largest possible number of band members?</question>

</context>

Please extract the information and question from the following sample: 

[ORIGINAL QUESTION]
0\end{lstlisting}

\newpage

\subsection{0-Shot Chain-of-Thought Prompts}\label{appx:cot_reasoning}
\begin{lstlisting}[
  caption={Prompt used for 0-shot chain-of-thought reasoning (0-CoT).},
  label={lst:default_0_cot_prompt}
]
[ORIGINAL QUESTION]

Please reason step by step, and put your final answer within \\boxed{}.
\end{lstlisting}

\begin{lstlisting}[
  caption={The shortened prompt used to elicit reasoning across the standard question.},
  label={lst:adjusted_0_cot_prompt}
]
"You are given ONLY the information in the following question. Use ONLY the facts provided by the question to compute the answer. 

[ORIGINAL QUESTION]

Please reason step by step, and put your final answer within \\boxed{}.
\end{lstlisting}

\subsection{Baseline Prompts}\label{appx:baseline_prompts}
\begin{lstlisting}[
  caption={Prompt used for Plan and Solve reasoning (PS).},
  label={lst:ps_prompt}
]
[ORIGINAL QUESTION]

Let's first understand the problem, extract relevant variables and their corresponding numerals, and make and devise a complete plan. Then, let's carry out the plan, calculate intermediate variables (pay attention to correct numerical calculation and commonsense), solve the problem step by step, and show the answer.

Please reason step by step, and put your final answer within \\boxed{}.
\end{lstlisting}

\begin{lstlisting}[
  caption={Prompt used for CoRe reasoning.},
  label={lst:core_prompt}
]
[ORIGINAL QUESTION]

Let's first understand the problem, then list all the known conditions, which are formed by numbers or quantitative relationships along with their contexts from the problem text, and identify the final goal of the problem.

Please reason step by step, and put your final answer within \\boxed{}.
\end{lstlisting}

\newpage

\subsection{F-CoT Reasoning with Context Blocks}\label{appx:context_reasoning}

In our F-CoT experiment, we used the following prompts shown in Listing~\ref{lst:reasoning_generation} to instruct the model to reason based on the provided context. The other prompts correspond to different variations from the sensitivity analysis.

\begin{lstlisting}[
  caption={The standard F-CoT prompt used to elicit reasoning across the context.},
  label={lst:reasoning_generation}
]
You are given ONLY the structured <context> below. Use ONLY the facts inside <context> to compute the answer. Do NOT reference or repeat the original natural-language question or any outside information.

Instructions:
1) Show step-by-step reasoning between <think> and </think> blocks.
2) In your reasoning, explicitly cite the relevant <info_k> entries when you use them, including the angle brackets (e.g., 'From <info_1>, ...').
3) After the reasoning, output the final answer in \\boxed{}.

<context>
[INSERTED CONTEXT]
</context>

Please reason step by step, and put your final answer within \\boxed{}.
\end{lstlisting}

\begin{lstlisting}[
  caption={The shortened prompt from the sensitivity analysis used to elicit reasoning across the context.},
  label={lst:reasoning_generation_short}
]
"You are given ONLY the structured <context> below. Use ONLY the facts inside <context> to compute the answer. 

<context>
[INSERTED CONTEXT]
</context>

Please reason step by step, and put your final answer within \\boxed{}.
\end{lstlisting}

\begin{lstlisting}[
  caption={Adjusted prompt from the sensitivity analysis for context variation 1.},
]
You are given ONLY the structured context below. Use ONLY the provided facts to compute the answer. Do NOT reference or repeat the original natural-language question or any outside information

Instructions:
1) Show step-by-step reasoning between <think> and </think> blocks.
2) In your reasoning, explicitly cite the relevant info_k entries when you use them (e.g., 'From info_1, ...').
3) After the reasoning, output the final answer in \\boxed{}.

[INSERTED CONTEXT - VARIATION 1]

Please reason step by step, and put your final answer within \\boxed{}.\end{lstlisting}

\newpage 
\begin{lstlisting}[
  caption={Adjusted prompt from the sensitivity analysis for context variation 2.},
]
You are given ONLY the structured information below. Use ONLY the provided facts to compute the answer. Do NOT reference or repeat the original natural-language question or any outside information.

Instructions:
1) Show step-by-step reasoning between <think> and </think> blocks.
2) In your reasoning, explicitly cite the relevant information when you use them.
3) After the reasoning, output the final answer in \\boxed{}.

[INSERTED CONTEXT - VARIATION 2]

Please reason step by step, and put your final answer within \\boxed{}.\end{lstlisting}

\begin{lstlisting}[
  caption={Adjusted prompt from the sensitivity analysis for context variation 3.},
]
You are given ONLY the structured information below. Use ONLY the provided facts to compute the answer. Do NOT reference or repeat the original natural-language question or any outside information.

Instructions:
1) Show step-by-step reasoning between <think> and </think> blocks.
2) After the reasoning, output the final answer in \\boxed{}.

[INSERTED CONTEXT - VARIATION 3]

Please reason step by step, and put your final answer within \\boxed{}.\end{lstlisting}

\begin{lstlisting}[
  caption={Adjusted prompt from the sensitivity analysis for context and original question.},
  label={lst:context_and_question}
]
You are given the structured <context> below and the original question. Use ONLY the facts inside <context> to compute the answer. Refer to the original question ONLY if some information in <context> is unclear or ambiguous.

Instructions:
1) Show step-by-step reasoning between <think> and </think> blocks.
2) In your reasoning, explicitly cite the relevant information when you use them.
3) After the reasoning, output the final answer in \\boxed{}.

Original question: [INSERTED QUESTION]

Context: [INSERTED CONTEXT]

Please reason step by step, and put your final answer within \\boxed{}.\end{lstlisting}

\newpage
\subsection{Context Designs}\label{appx:context_designs}
The context design in \cref{lst:default_context} corresponds to the default context variant. The other context designs are explored in our sensitivity analysis.

\begin{lstlisting}[
  caption={Standard F-CoT context with XML-like format.},
  label={lst:default_context}
]
<context>
    <info_1>Price per apple: $2<info_1>
    <info_2>Price per pear: 1.5x price per apple<info_2>
    <question>Total cost of 2 apples and 2 pears</question>
</context>
\end{lstlisting}

\begin{lstlisting}[
  caption={Context alternative design 1 - enumerated list.},
  label={lst:context_variation_1}
]
Context:
info_1: Price per apple: $2
info_2: Price per pear: 1.5x price per apple
question: Total cost of 2 apples and 2 pears</question>
\end{lstlisting}

\begin{lstlisting}[
  caption={Context alternative design 2 - unnumbered list.},
  label={lst:context_variation_2}
]
**Information**
- Price per apple: $2
- Price per pear: 1.5x price per apple

**Question**
Total cost of 2 apples and 2 pears</question>
\end{lstlisting}

\begin{lstlisting}[
  caption={Context alternative design 3 - concatenated information.},
  label={lst:context_variation_3}
]
Price per apple: $2. Price per pear: 1.5x price per apple. Question: Total cost of 2 apples and 2 pears</question>
\end{lstlisting}

\newpage

\subsection{Sentence Type Annotation}

\begin{lstlisting}[
  caption={Prompt used annotating the sentence type.},
]

You are an annotator distinguishing between *extraction* and *reasoning* in model-generated solutions.  

- **Extraction** = A sentence that copies or paraphrases information directly from the input question without transformation or inference.  

- **Reasoning** = A sentence that performs logical inference, arithmetic, explanation, or deduction not explicitly given in the input.  

- **Filler** = A sentence that does not contain extraction or reasoning, such as generic phrases ("Let's solve step by step", "Now I will calculate").  

Instructions:

1. I will give you a question and a step-by-step solution.  

2. Split the solution into individual sentences.  

3. For each sentence, output one of the following labels: `Extraction`, `Reasoning`, or `Filler`.  

4. Return the result as a JSON list of objects, where each object has two keys: `"sentence"` and `"label"`.  

---

Question: [ORIGINAL QUESTION] 

Solution: [MODEL GENERATION]

Output format:

[

{"sentence": "...", "label": "..."},

{"sentence": "...", "label": "..."}

]

\end{lstlisting}

\newpage
\subsection{Overthinking Score}\label{appx:overthinking_score}

\begin{lstlisting}[
  caption={The prompt used to compute the overthinking score. Model: GPT-5 Mini (August 2025 version).},
  label={lst:prompt_overthinking_score}
]

You are an AI judge focused on detecting when a model prefers its internal reasoning chain over performing concrete mathematical reasoning steps that advance toward a solution.

<INTERACTION>

User Question: [ORIGINAL QUESTION]

Model Solution: [MODEL GENERATION]

</INTERACTION>

Analyze the <INTERACTION> and determine if the model is preferring internal reasoning and reflection over actually performing concrete steps to solve the problem.

How could this be detected?

<CORE PRINCIPLE>
- Analysis Paralysis:
  The model repeatedly analyzes strategies or discusses what could be done (e.g., substitution, factoring, induction) without executing any mathematical steps.
- Rogue Reasoning:
  The model jumps between multiple methods (e.g., algebraic -> geometric -> numerical) without completing any of them or checking intermediate results.
- Premature Disengagement:
  The model concludes the problem without verifying or deriving the answer, often due to overconfidence ("The answer must be X") or self-dismissal ("This can't be solved").
</CORE PRINCIPLE>

<SCORING SYSTEM (0-10)>
0-3: Clear, step-by-step reasoning
- Progresses directly toward the solution through algebraic, numerical, or logical steps.
- A brief plan or method selection is fine if quickly followed by computation.
- Checks and verifies results.
- Repeats calculations or re-derives expressions carefully.
- Stays focused on the mathematical task.

4-7: Some overthinking but still problem-solving
- Includes long reasoning or multiple reflections before computing, but eventually performs concrete steps.
- Might discuss multiple solution paths once but settles on one.
- Verification is present but delayed.
- Shows minor hesitation or digression but remains mostly focused.

8-10: Strong overthinking / minimal problem interaction
- Engages in extended speculation about possible methods or properties without any real computation.
- Switches between multiple unrelated strategies.
- Concludes without a derivation.
- Produces long internal dialogue instead of concrete reasoning steps.
- Gets stuck in unproductive self-reflection.
</SCORING SYSTEM>

<ANALYSIS STEPS>
1. Analysis Paralysis
   - Does the model spend a long time discussing potential approaches?
   - Are there few or no concrete equations, simplifications, or calculations?

2. Rogue Reasoning
   - Does it jump between many approaches without applying or finishing any?
   - Are intermediate results ignored or overwritten by new plans?

3. Premature Disengagement
   - Does it state a final answer without derivation or checking?
   - Does it give up or claim it cannot solve the problem?
</ANALYSIS STEPS>

<EXAMPLES>
Example 1 - Focused Reasoning (Good):
Problem: Solve x^2 - 5x + 6 = 0.
Model:
"Let's compute the discriminant: D = 25 - 24 = 1.
Then x = (5 \pm 1)/2, so x = 2 or x = 3."
Score: 0 - The model works step-by-step, no unnecessary reflection.

Example 2 - Brief Planning (Good):
Model:
"This looks quadratic, so I can either factor or use the quadratic formula. Let's try factoring: (x-2)(x-3) = 0, giving roots 2 and 3."
Score: 2 - Some internal planning, but quickly applied and solved.

Example 3 - Mild Overthinking:
Model:
"We might approach this by factoring, or perhaps by completing the square. Factoring works if the product and sum fit correctly... Hmm, 2×3 = 6 and 2+3 = 5, that works. So, yes, x=2,3."
Score: 5 - Some unnecessary deliberation, but correct and eventually concrete.

Example 4 - Analysis Paralysis:
Model:
"This problem could be tackled in multiple ways - factoring, completing the square, or graphing. Maybe I should first recall the discriminant formula, but perhaps the sum-product relation is easier. Alternatively, we might think geometrically about the roots of a parabola... On the other hand..."
Score: 9 - Prolonged reflection with no actual computation.

Example 5 - Rogue Reasoning:
Model:
"Let's test some integer values: x=0 gives 6, x=1 gives 2, x=2 gives 0 so that works. But maybe we can confirm using the discriminant method... or perhaps symmetry arguments from the vertex form... though also we might estimate roots by graph intersection."
Score: 8 - Multiple scattered strategies without systematic follow-through.

Example 6 - Premature Disengagement:
Model:
"The answer should probably be around 2 or 3 because it looks symmetric."
Score: 10 - Concludes without derivation or verification.
</EXAMPLES>

<IMPORTANT>
Format your response as:
<answer>
{

"overthinking_score": "[0-10]",
"reasoning": "Explain your reasoning for the score, in one line suitable for JSON."
}
</answer>

Always surround your answer with <answer> and </answer> tags.
Take your time to understand the reasoning trace and analyze it carefully.
Think step-by-step about whether the model overthinks (reflects excessively) instead of reasoning concretely toward a mathematical solution.
</IMPORTANT>

\end{lstlisting}

%% file: sections/B_qualitative_examples.tex
\clearpage
\section{Qualitative Results}

\subsection{GSM-Hard Results}\label{appx:gsm_hard_comparison}
\begin{lstlisting}[
  caption={Generated output for Qwen3 14B queried with a GSM-Hard Example and 0-CoT. The total number of generated tokens is 6,401. Parts of the output are shortened, indicated by ... .},
]
<think>
Okay, let's see. Janet has ducks that lay 16 eggs each day. She eats three for breakfast every morning. Then she bakes muffins using 4933828... wait, that number seems really big. Maybe that's a typo? Let me check the problem again.

"Janet's ducks lay 16 eggs per day. She eats three for breakfast every morning and bakes muffins for her friends every day with 4933828. She sells the remainder at the farmers' market daily for $2 per fresh duck egg. How much in dollars does she make every day at the farmers' market?"

Hmm, baking muffins with 4933828... that number is way too large. Maybe it's supposed to be 493,382.8? Or maybe 493.3828? Or perhaps it's a different number? Wait, maybe it's a typo and the number is supposed to be 493, but written as 4933828? Let me think. If the number is 493,382.8, that would be like 493 thousand eggs, which is way more than the 16 eggs she has. That doesn't make sense. If she only has 16 eggs a day, she can't use 493 thousand eggs to bake muffins. So maybe there's a mistake in the number? Maybe it's 4.933828? Or maybe it's 49 eggs? Wait, the original problem says "bakes muffins for her friends every day with 4933828." Maybe that's supposed to be 49 eggs? Let me check again.

Wait, the original problem says "bakes muffins for her friends every day with 4933828." Maybe that's a typo and should be 49 eggs? Or maybe 4.933828? But the number is written as 4933828. Maybe it's a misinput. Let me check if that number is related to something else. Alternatively, maybe the number is correct, but it's in a different unit? Wait, that seems unlikely. 

[...]

But given that the original number is 4933828, which is 4,933,828, and if it's a misinput of 4.933828, then the answer is approximately $16.13. 

However, since the problem might have a different intended number, like 493,382.8 divided by 100,000, which is 4.933828, I think that's the best I can do. 

So, I'll go with $16.13 as the answer, but I'm not fully confident due to the ambiguity in the number provided.

**Final Answer**
\boxed{16.13}
\end{lstlisting}

\clearpage
\begin{lstlisting}[
  caption={Generated output for Qwen3 14B queried with a GSM-Hard Example and F-CoT. The total number of generated tokens is 2,820. Parts of the output are shortened, indicated by ... .},
]
Okay, let's see. The problem is asking how much money she makes per day at the farmers' market. The context gives me some information about eggs.

First, from info_1, ducks lay 16 eggs per day. So total eggs available each day are 16.

Then, info_2 says she eats 3 eggs for breakfast every morning. So that's 3 eggs used up by her.

Next, info_3 mentions she uses 4,933,828 eggs daily to bake muffins. Wait, that number seems really large. Let me check. Oh, maybe that's a typo? Because if ducks only lay 16 eggs a day, how can she use over 4 million eggs for muffins? That doesn't make sense. But the problem says to use only the given facts. Maybe there's a mistake in the numbers? Or maybe the numbers are correct? Wait, maybe the info_3 is a typo. Let me think again. If the numbers are as given, then perhaps the problem is designed with those numbers. But that would mean that the total eggs used would be 3 (breakfast) + 4,933,828 (muffins) = 4,933,831 eggs. But the total eggs laid per day are only 16. So that would mean she's using way more eggs than are available. That's impossible. But since the problem says to use only the given facts, maybe I need to proceed regardless. Wait, maybe the info_3 is a mistake, like maybe it's supposed to be 4.933828 eggs? But the number is written as 4,933,828, which is 4 million. Hmm. Alternatively, maybe the problem is designed to have a negative number of eggs sold? But that would be odd. Alternatively, maybe the numbers are correct and the answer is that she can't sell any eggs because she uses more than she has. But the problem says to compute the answer based on the given info. Let me check again.

[...]

This is confusing. But the user says to use only the facts inside the context. So I need to proceed with the numbers given. 

[...]

This is a tricky situation. Given the instructions, I have to use the given numbers, even if they are inconsistent. So the answer would be the calculation based on the numbers, even if it's negative. Therefore, the final answer would be \boxed{-9867630}. But I'm not sure if that's what the problem expects. However, given the strict instructions, this is the only possible answer.
</think>

[...]
### Final Answer
$$
\boxed{-9867630}
$$
\end{lstlisting}

\clearpage

%% file: sections/C_quantitative_results.tex
\section{Quantitative Results}

\subsection{Experiments with GPT-5 Mini Pre-Computed Context Information}\label{appx:precomputed_context_results}

\cref{tab:results_precomputed} summarizes the performance of multiple reasoning LLMs across five benchmarks, comparing standard zero-shot CoT (0-CoT) with our F-CoT approach, as well as Plan-and-Solve (PS) and CoRe prompting strategies. For F-CoT, the context is pre-computed using GPT-5 Mini, ensuring structured information is provided to the model before reasoning.

\begin{table}[ht]
    \setlength{\tabcolsep}{3pt}
    \centering
    \caption{Performance of reasoning LLMs on five benchmarks using context pre-computed by GPT-5 Mini.}
    \vspace{-0.1cm}
    \resizebox{\columnwidth}{!}{
        \begin{tabular}{llccccccccccc}
        \toprule
        \textbf{Model} & \textbf{Prompting} & \multicolumn{2}{c}{\textbf{SVAMP}} & \multicolumn{2}{c}{\textbf{GSM-Hard}} & \multicolumn{2}{c}{\textbf{MATH-500}} & \multicolumn{2}{c}{\textbf{AIME2024}} & \multicolumn{2}{c}{\textbf{AIME2025}} &  \\
        & & Pass@5 & \# Tokens & Pass@5 & \# Tokens & Pass@5 & \# Tokens & Pass@5 & \# Tokens & Pass@5 & \# Tokens \\
        \midrule
        \multirow{4}{*}{\textbf{Qwen 3 0.6B}} & 0-CoT & 95.67\% & 1,358 & 65.88\% & 5,235 & 90.20\% & 5,643 & 30.00\% & 16,907 & 23.33\% & 14,904  \\
         & PS & 96.00\% & 1,397 & 65.43\% & 4,989 & 88.40\% & 5,522 & 30.00\% & 16,653 & 26.67\% & 14,289 \\
         & CoRe & 96.33\% & 1,380 & 66.34\% & 4,854 & 91.00\% & 5,140 & 16.67\% & 17,833 & 23.33\% & 14,035 \\
         & F-CoT & 94.33\% & 434 & 64.90\% & 2,122 & 87.20\% & 2,665 & 26.67\% & 9,615 & 23.33\% & 8,460 \\
         \midrule
        \multirow{4}{*}{\textbf{Qwen 3 4B}} & 0-CoT & 97.00\% & 1,871 & 70.36\% & 5,674 & 98.80\% & 5,099 & 86.67\% & 12,640 & 83.33\% & 16,293 \\
         & PS & 96.67\% & 1,706 & 71.49\% & 5,318 & 98.60\% & 5,100 & 83.33\% & 12,924 & 80.00\% & 16,115 \\
         & CoRe & 97.00\% & 1,844 & 71.95\% & 5,509 & 98.40\% & 5,005 & 83.33\% & 13,541 & 76.67\% & 16,695 \\
         & F-CoT & 95.00\% & 535 & 76.88\% & 2,308 & 98.40\% & 2,580 & 83.33\% & 9,422 & 73.33\% & 11,287 \\
         \midrule
        \multirow{4}{*}{\textbf{Qwen 3 14B}} & 0-CoT & 97.00\% & 1,487 & 71.87\% & 4,429 & 99.40\% & 4,931 & 86.67\% & 12,697 & 80.00\% & 16,185 \\
         & PS & 97.00\% & 1,457 & 72.86\% & 4,285 & 99.40\% & 4,697 & 90.00\% & 13,276 & 83.33\% & 15,701 \\
         & CoRe & 96.67\% & 1,481 & 73.31\% & 4,253 & 98.80\% & 4,692 & 90.00\% & 14,255 & 80.00\% & 16,086 \\
        & F-CoT & 95.67\% & 476 & 77.03\% & 1,838 & 98.60\% & 2,437 & 83.33\% & 10,593 & 76.67\% & 12,788 \\
        \midrule
        \multirow{4}{*}{\textbf{Qwen 3 32B}} & 0-CoT & 97.00\% & 1,377 & 70.13\% & 4,516 & 99.00\% & 4,384 & 90.00\% & 12,343 & 86.67\% & 15,137 \\
         & PS & 96.00\% & 1,395 & 70.96\% & 4,238 & 99.00\% & 4,397 & 90.00\% & 12,358 & 86.67\% & 14,644 \\
         & CoRe & 96.33\% & 1,464 & 71.27\% & 4,387 & 99.20\% & 4,688 & 90.00\% & 12,505 & 83.33\% & 15,456\\
         & F-CoT & 95.67\% & 429 & 77.03\% & 1,785 & 99.00\% & 2,309 & 83.33\% & 8,964 & 76.67\% & 11,143 \\
        \bottomrule 
        \end{tabular}
        }
        \label{tab:results_precomputed}
\end{table}

\subsection{Self-Computed Context Information}\label{appx:self_generated_context_results}

\cref{tab:results_self_computed_context} summarizes the performance of multiple reasoning LLMs across five benchmarks, comparing F-CoT using self-generated context with F-CoT using context pre-computed by GPT-5 Mini.

\begin{table}[ht]
    \setlength{\tabcolsep}{3pt}
    \centering
    \caption{Performance of reasoning LLMs on five benchmarks using context computed by the models themselves.}
    \vspace{-0.1cm}
    \resizebox{\columnwidth}{!}{
        \begin{tabular}{ll|ccc|ccc|ccc|ccc|cccc}
        \toprule
        \textbf{Model} & \textbf{Prompting} & \multicolumn{3}{c}{\textbf{SVAMP}} & \multicolumn{3}{c}{\textbf{GSM-Hard}} & \multicolumn{3}{c}{\textbf{MATH-500}} & \multicolumn{3}{c}{\textbf{AIME2024}} & \multicolumn{3}{c}{\textbf{AIME2025}} &  \\
        & & Pass@5 & \# Tokens & \makecell{Valid\\Context} & Pass@5 & \# Tokens & \makecell{Valid\\Context} & Pass@5 & \# Tokens & \makecell{Valid\\Context} & Pass@5 & \# Tokens & \makecell{Valid\\Context} & Pass@5 & \# Tokens & \makecell{Valid\\Context} \\
        \midrule
        \multirow{3}{*}{\textbf{Qwen 3 0.6B}} & 0-CoT & 95.67\% & 1,358 & N/A & 65.88\% & 5,235 & N/A & 90.20\% & 5,643 & N/A & 30.00\% & 16,907 & N/A & 23.33\% & 14,904 & N/A  \\
         & F-CoT (Provided) & 94.33\% & 434 & 100\% & 64.90\% & 2,122 & 100\% & 87.20\% & 2,665 & 100\% & 26.67\% & 9,615 & 100\% & 23.33\% & 8,460 & 100\% \\
         & F-CoT (Self-Generated) & 64.00\% & 466 & 0.00\% & 22.37\% & 934 & 0.01\% & 24.40\% & 1,055 & 1.20\% & 3.33\% & 2,851 & 3.33\% & 3.33\% &  2,087 & 0.00\%  \\
         \midrule
        \multirow{3}{*}{\textbf{Qwen 3 4B}} & 0-CoT & 97.00\% & 1,871 & N/A & 70.36\% & 5,674 & N/A & 98.80\% & 5,099 & N/A & 86.67\% & 12,640 & N/A & 83.33\% & 16,293 & N/A \\
         & F-CoT (Provided) & 95.00\% & 535 & 100\% & 76.88\% & 2,308 & 100\% & 98.40\% & 2,580 & 100\% & 83.33\% & 9,422 & 100\% & 73.33\% & 11,287 & 100\% \\
         & F-CoT (Self-Generated) & 91.00\% & 786 & 98.67\% & 66.11\% & 2,403 & 98.94\% & 93.80 & 2,990 & 99.80\% & 60.00\% & 9,253 & 100\% & 66.67\% & 12,439 & 100.0\%  \\
         \midrule
        \multirow{3}{*}{\textbf{Qwen 3 14B}} & 0-CoT & 97.00\% & 1,487 & N/A & 71.87\% & 4,429 & N/A & 99.40\% & 4,931 & N/A & 86.67\% & 12,697 & N/A & 80.00\% & 16,185 & N/A \\
        & F-CoT (Provided) & 95.67\% & 476 & 100\% & 77.03\% & 1,838 & 100\% & 98.60\% & 2,437 & 100\% & 83.33\% & 10,593 & 100\% & 76.67\% & 12,788 & 100\% \\
        & F-CoT (Self-Generated) & 94.00\% & 581 & 99.67\% & 70.51\% & 1,963 & 99.55\% & 98.00\% & 2,526 & 100\% & 80.00\% & 9,389 & 100\% & 76.67\% & 11,695 & 100\% \\
        \midrule
        \multirow{3}{*}{\textbf{Qwen 3 32B}} & 0-CoT & 97.00\% & 1,377 & N/A & 70.13\% & 4,516 & N/A & 99.00\% & 4,384 & N/A & 90.00\% & 12,343 & N/A & 86.67\% & 15,137 & N/A \\
         & F-CoT (Provided) & 95.67\% & 429 & 100\% & 77.03\% & 1,785 & 100\% & 99.00\% & 2,309 & 100\% & 83.33\% & 8,964 & 100\% & 76.67\% & 11,143 & 100\% \\
         & F-CoT (Self-Generated) & 95.67\% & 556 & 100\% & 73.77\% & 1,827 & 99.92\% & 98.00\% & 2,547 & 100\% & 86.67\% & 9,395 & 100\% & 80.00\% & 11,040 & 100\% \\
        \bottomrule 
        \end{tabular}
        }
        \label{tab:results_self_computed_context}
\end{table}

\newpage
\subsection{Ablation Study and Sensitivity Analysis}\label{appx:sensitivity_results}
We conduct a series of ablation and sensitivity experiments to assess the robustness of F-CoT under various prompt and context conditions. \cref{tab:sensitivity_results} shows the effect of different reasoning prompt designs and context representations on the Qwen3 14B model using GPT-5 Mini pre-computed context. Importantly, the information and sentences between the various formats are identical, and only their representation (XML, list, concatenated) is changed.

We observe that while the default F-CoT prompt achieves a substantial reduction in token count compared to 0-CoT, adjusting the prompt or including the original question alongside the pre-computed context can slightly improve accuracy with minimal impact on efficiency. Alternative context formats (enumerated list, unnumbered list, or concatenated text) produce similar accuracy, confirming that F-CoT is largely robust to the exact structure of the extracted information.

\cref{tab:sensitivity_results_icl} investigates the effect of in-context examples (ICL) for self-generated context extraction on Qwen3 14B. Including few-shot examples has a negligible impact on Pass@5 and token efficiency, though it slightly decreases the proportion of valid context blocks.

Finally, \cref{tab:sensitivity_results_model_size} evaluates how model size affects context quality and reasoning performance. Smaller models (0.6B) struggle to produce valid contexts, resulting in a dramatic drop in accuracy. Larger models consistently generate valid contexts and achieve high Pass@5 scores, highlighting the benefit of using more capable LLMs for context extraction in F-CoT. Overall, these analyses demonstrate that F-CoT is robust across prompt variations, context formats, and model scales, while retaining its efficiency advantages.

\begin{table}[ht]
    \setlength{\tabcolsep}{3pt}
    \centering
    \caption{Sensitivity Analysis performed on Qwen3 14B with GPT5-mini precomputed context}
    \vspace{-0.1cm}
        \begin{tabular}{lccc}
        \toprule
        \makecell{\textbf{Setting}} & \multicolumn{2}{c}{\textbf{MATH-500}} \\
        & Pass@5 & \# Tokens \\
        \midrule
        0-CoT  -- Default Prompt (\cref{lst:default_0_cot_prompt}) & 99.40\% & 4,931 \\
        0-CoT -- Adjusted Prompt (\cref{lst:adjusted_0_cot_prompt}) & 99.20\% & 4,246 \\
        \midrule
        F-CoT -- Default Prompt (\cref{lst:reasoning_generation}) & 98.60\% & 2,437 \\
        F-CoT -- Adjusted Prompt (\cref{lst:reasoning_generation_short})  & 99.20\% & 3,184 \\
        \midrule
        F-CoT -- Context Variation 1 (\cref{lst:context_variation_1}) & 98.80\% & 2,453 \\
        F-CoT -- Context Variation 2 (\cref{lst:context_variation_2}) & 98.80\% & 2,744 \\
        F-CoT -- Context Variation 3 (\cref{lst:context_variation_3}) & 98.40\% & 2,707 \\
        \midrule
        F-CoT -- Context + Original Question & 99.20\% & 2,898 \\
        \bottomrule 
        \end{tabular}
    \label{tab:sensitivity_results}
\end{table}

\begin{table}[ht]
    \setlength{\tabcolsep}{3pt}
    \centering
    \caption{Sensitivity Analysis Context Extraction Prompt on Qwen-14B (Self-generated context)}
    \vspace{-0.1cm}
        \begin{tabular}{lcccc}
        \toprule
        \makecell{\textbf{Setting}} & \multicolumn{3}{c}{\textbf{MATH-500}} \\
                                    & Pass@5    & \# Tokens & \makecell{Valid\\Context} \\
        \midrule
        F-CoT  -- Prompt w/o ICL (\cref{lst:context_extraction})    & 97.00\% & 2,526 & 100\% \\
        F-CoT -- Prompt w/ ICL (\cref{lst:prompt_math_extraction})  & 96.80\% & 2,563 & 98.20\%\\
        \bottomrule 
        \end{tabular}
    \label{tab:sensitivity_results_icl}
\end{table}

\begin{table}[ht]
    \setlength{\tabcolsep}{3pt}
    \centering
    \caption{Sensitivity Analysis for Context Extraction with Models of Different Sizes on Qwen-14B.}
    \vspace{-0.1cm}
        \begin{tabular}{lcccc}
        \toprule
        \makecell{\textbf{Setting}} & \multicolumn{3}{c}{\textbf{MATH-500}} \\
                                    & Pass@5    & \# Tokens & \makecell{Valid\\Context} \\
        \midrule
        F-CoT  -- Qwen3-0.6B Context & 41.2\% & 1,472 & 0.2\% \\
        F-CoT -- Qwen3-1.7B Context  & 91.4\% & 2,378 & 95.2\% \\
        F-CoT -- Qwen3-4B Context    & 92.8\% & 2,496 & 99.8\%\\
        F-CoT -- Qwen3-8B Context    & 94.8\% & 2,547 & 99.6\% \\
        F-CoT -- Qwen3-14B Context   & 97.2\% & 2,625 & 99.8\%\\
        F-CoT -- Qwen3-32B Context   & 97.0\% & 2,542 & 99.8\%\\
        \bottomrule 
        \end{tabular}
    \label{tab:sensitivity_results_model_size}
\end{table}

%% file: references.bib
@inproceedings{coleg_xu,
    title={Can {LLM}s Solve Longer Math Word Problems Better?},
    author={Xin Xu and Tong Xiao and Zitong Chao and Zhenya Huang and Can Yang and Yang Wang},
    booktitle={The Thirteenth International Conference on Learning Representations},
    year={2025},
}

@article{cobbe2021gsm8k,
  title={Training Verifiers to Solve Math Word Problems},
  author={Cobbe, Karl and Kosaraju, Vineet and Bavarian, Mohammad and Chen, Mark and Jun, Heewoo and Kaiser, Lukasz and Plappert, Matthias and Tworek, Jerry and Hilton, Jacob and Nakano, Reiichiro and Hesse, Christopher and Schulman, John},
  journal={arXiv preprint arXiv:2110.14168},
  year={2021}
}

@article{wei2022chain,
  title={Chain-of-thought prompting elicits reasoning in large language models},
  author={Wei, Jason and Wang, Xuezhi and Schuurmans, Dale and Bosma, Maarten and Xia, Fei and Chi, Ed and Le, Quoc V and Zhou, Denny and others},
  journal={Advances in neural information processing systems},
  volume={35},
  pages={24824--24837},
  year={2022}
}

@article{kojima2022large,
  title={Large language models are zero-shot reasoners},
  author={Kojima, Takeshi and Gu, Shixiang Shane and Reid, Machel and Matsuo, Yutaka and Iwasawa, Yusuke},
  journal={Advances in neural information processing systems},
  volume={35},
  pages={22199--22213},
  year={2022}
}

@article{wang2022self,
  title={Self-instruct: Aligning language models with self-generated instructions},
  author={Wang, Yizhong and Kordi, Yeganeh and Mishra, Swaroop and Liu, Alisa and Smith, Noah A and Khashabi, Daniel and Hajishirzi, Hannaneh},
  journal={arXiv preprint arXiv:2212.10560},
  year={2022}
}

@article{yao2023tree,
  title={Tree of thoughts: Deliberate problem solving with large language models},
  author={Yao, Shunyu and Yu, Dian and Zhao, Jeffrey and Shafran, Izhak and Griffiths, Tom and Cao, Yuan and Narasimhan, Karthik},
  journal={Advances in neural information processing systems},
  volume={36},
  pages={11809--11822},
  year={2023}
}

@inproceedings{yao2023react,
  title={React: Synergizing reasoning and acting in language models},
  author={Yao, Shunyu and Zhao, Jeffrey and Yu, Dian and Du, Nan and Shafran, Izhak and Narasimhan, Karthik and Cao, Yuan},
  booktitle={International Conference on Learning Representations (ICLR)},
  year={2023}
}

@article{kahneman2011thinking,
  title={Thinking, fast and slow},
  author={Kahneman, Daniel},
  journal={Farrar, Straus and Giroux},
  year={2011}
}

@misc{deepseekai2025deepseekr1incentivizingreasoningcapability,
      title={DeepSeek-R1: Incentivizing Reasoning Capability in LLMs via Reinforcement Learning}, 
      author={DeepSeek-AI},
      year={2025},
      eprint={2501.12948},
      archivePrefix={arXiv},
      primaryClass={cs.CL},
}

@misc{qwen3technicalreport,
      title={Qwen3 Technical Report}, 
      author={{Qwen Team}},
      year={2025},
      eprint={2505.09388},
      archivePrefix={arXiv},
      primaryClass={cs.CL},
}

@article{muennighoff2025s1,
  title={s1: Simple test-time scaling},
  author={Muennighoff, Niklas and Yang, Zitong and Shi, Weijia and Li, Xiang Lisa and Fei-Fei, Li and Hajishirzi, Hannaneh and Zettlemoyer, Luke and Liang, Percy and Cand{\`e}s, Emmanuel and Hashimoto, Tatsunori},
  journal={arXiv preprint arXiv:2501.19393},
  year={2025}
}

@article{madaan2023self,
  title={Self-refine: Iterative refinement with self-feedback},
  author={Madaan, Aman and Tandon, Niket and Gupta, Prakhar and Hallinan, Skyler and Gao, Luyu and Wiegreffe, Sarah and Alon, Uri and Dziri, Nouha and Prabhumoye, Shrimai and Yang, Yiming and others},
  journal={Advances in Neural Information Processing Systems},
  volume={36},
  pages={46534--46594},
  year={2023}
}

@article{tian2025think,
  title={Think twice: Enhancing llm reasoning by scaling multi-round test-time thinking},
  author={Tian, Xiaoyu and Zhao, Sitong and Wang, Haotian and Chen, Shuaiting and Ji, Yunjie and Peng, Yiping and Zhao, Han and Li, Xiangang},
  journal={arXiv preprint arXiv:2503.19855},
  year={2025}
}

@article{aggarwal2025l1,
  title={L1: Controlling how long a reasoning model thinks with reinforcement learning},
  author={Aggarwal, Pranjal and Welleck, Sean},
  journal={arXiv preprint arXiv:2503.04697},
  year={2025}
}

@inproceedings{chen2024not,
title={Do {NOT} Think That Much for 2+3=? On the Overthinking of Long Reasoning Models},
author={Xingyu Chen and Jiahao Xu and Tian Liang and Zhiwei He and Jianhui Pang and Dian Yu and Linfeng Song and Qiuzhi Liu and Mengfei Zhou and Zhuosheng Zhang and Rui Wang and Zhaopeng Tu and Haitao Mi and Dong Yu},
booktitle={Forty-second International Conference on Machine Learning},
year={2025},
}

@article{yeo2025demystifying,
  title={Demystifying long chain-of-thought reasoning in llms},
  author={Yeo, Edward and Tong, Yuxuan and Niu, Morry and Neubig, Graham and Yue, Xiang},
  journal={arXiv preprint arXiv:2502.03373},
  year={2025}
}

@article{arora2025training,
  title={Training language models to reason efficiently},
  author={Arora, Daman and Zanette, Andrea},
  journal={arXiv preprint arXiv:2502.04463},
  year={2025}
}

@article{ye2025emergence,
  title={On the emergence of thinking in llms i: Searching for the right intuition},
  author={Ye, Guanghao and Pham, Khiem Duc and Zhang, Xinzhi and Gopi, Sivakanth and Peng, Baolin and Li, Beibin and Kulkarni, Janardhan and Inan, Huseyin A},
  journal={arXiv preprint arXiv:2502.06773},
  year={2025}
}

@article{cuadron2025danger,
  title={The danger of overthinking: Examining the reasoning-action dilemma in agentic tasks},
  author={Cuadron, Alejandro and Li, Dacheng and Ma, Wenjie and Wang, Xingyao and Wang, Yichuan and Zhuang, Siyuan and Liu, Shu and Schroeder, Luis Gaspar and Xia, Tian and Mao, Huanzhi and others},
  journal={arXiv preprint arXiv:2502.08235},
  year={2025}
}

@article{zhuang2025accelerating,
  title={Accelerating chain-of-thought reasoning: When goal-gradient importance meets dynamic skipping},
  author={Zhuang, Ren and Wang, Ben and Sun, Shuifa},
  journal={arXiv preprint arXiv:2505.08392},
  year={2025}
}

@article{xia2025tokenskip,
  title={Tokenskip: Controllable chain-of-thought compression in llms},
  author={Xia, Heming and Leong, Chak Tou and Wang, Wenjie and Li, Yongqi and Li, Wenjie},
  journal={arXiv preprint arXiv:2502.12067},
  year={2025}
}

@article{yuan2025not,
  title={Not All Tokens Are What You Need In Thinking},
  author={Yuan, Hang and Yu, Bin and Li, Haotian and Yang, Shijun and Wang, Christina Dan and Yu, Zhou and Xu, Xueyin and Qi, Weizhen and Chen, Kai},
  journal={arXiv preprint arXiv:2505.17827},
  year={2025}
}

@article{xiao2025limopro,
  title={LIMOPro: Reasoning Refinement for Efficient and Effective Test-time Scaling},
  author={Xiao, Yang and Wang, Jiashuo and Yuan, Ruifeng and Xu, Chunpu and Xu, Kaishuai and Li, Wenjie and Liu, Pengfei},
  journal={arXiv preprint arXiv:2505.19187},
  year={2025}
}

@article{yu2025long,
  title={Long-short chain-of-thought mixture supervised fine-tuning eliciting efficient reasoning in large language models},
  author={Yu, Bin and Yuan, Hang and Li, Haotian and Xu, Xueyin and Wei, Yuliang and Wang, Bailing and Qi, Weizhen and Chen, Kai},
  journal={arXiv preprint arXiv:2505.03469},
  year={2025}
}

@article{luo2025autol2s,
  title={Autol2s: Auto long-short reasoning for efficient large language models},
  author={Luo, Feng and Chuang, Yu-Neng and Wang, Guanchu and Le, Hoang Anh Duy and Zhong, Shaochen and Liu, Hongyi and Yuan, Jiayi and Sui, Yang and Braverman, Vladimir and Chaudhary, Vipin and others},
  journal={arXiv preprint arXiv:2505.22662},
  year={2025}
}

@article{dubey2024llama,
  title={The llama 3 herd of models},
  author={Dubey, Abhimanyu and Jauhri, Abhinav and Pandey, Abhinav and Kadian, Abhishek and Al-Dahle, Ahmad and Letman, Aiesha and Mathur, Akhil and Schelten, Alan and Yang, Amy and Fan, Angela and others},
  journal={arXiv e-prints},
  pages={arXiv--2407},
  year={2024}
}

@article{hendrycksmath2021,
  title={Measuring Mathematical Problem Solving With the MATH Dataset},
  author={Dan Hendrycks and Collin Burns and Saurav Kadavath and Akul Arora and Steven Basart and Eric Tang and Dawn Song and Jacob Steinhardt},
  journal={NeurIPS},
  year={2021}
}

@article{wang2023plan,
  title={Plan-and-solve prompting: Improving zero-shot chain-of-thought reasoning by large language models},
  author={Wang, Lei and Xu, Wanyu and Lan, Yihuai and Hu, Zhiqiang and Lan, Yunshi and Lee, Roy Ka-Wei and Lim, Ee-Peng},
  journal={arXiv preprint arXiv:2305.04091},
  year={2023}
}

@inproceedings{gao2023pal,
  title={Pal: Program-aided language models},
  author={Gao, Luyu and Madaan, Aman and Zhou, Shuyan and Alon, Uri and Liu, Pengfei and Yang, Yiming and Callan, Jamie and Neubig, Graham},
  booktitle={International Conference on Machine Learning},
  pages={10764--10799},
  year={2023},
  organization={PMLR}
}

@article{patel2021nlp,
  title={Are NLP models really able to solve simple math word problems?},
  author={Patel, Arkil and Bhattamishra, Satwik and Goyal, Navin},
  journal={arXiv preprint arXiv:2103.07191},
  year={2021}
}

@mist{aime2025,
  title={AIME Problems and Solutions},
  author={American Mathematics Competitions},
  year={2025},
  note={\url{https://artofproblemsolving.com/wiki/index.php/AIME_Problems_and_Solutions}}
}

@article{golchin2023time,
  title={Time travel in llms: Tracing data contamination in large language models},
  author={Golchin, Shahriar and Surdeanu, Mihai},
  journal={arXiv preprint arXiv:2308.08493},
  year={2023}
}

@inproceedings{oren2023proving,
  title={Proving test set contamination in black-box language models},
  author={Oren, Yonatan and Meister, Nicole and Chatterji, Niladri S and Ladhak, Faisal and Hashimoto, Tatsunori},
  booktitle={The Twelfth International Conference on Learning Representations},
  year={2023}
}

@inproceedings{sclar2024quantifying,
title={Quantifying Language Models' Sensitivity to Spurious Features in Prompt Design or: How I learned to start worrying about prompt formatting},
author={Melanie Sclar and Yejin Choi and Yulia Tsvetkov and Alane Suhr},
booktitle={The Twelfth International Conference on Learning Representations},
year={2024},
}

@inproceedings{errica2025did,
  title={What Did I Do Wrong? Quantifying LLMs’ Sensitivity and Consistency to Prompt Engineering},
  author={Errica, Federico and Sanvito, Davide and Siracusano, Giuseppe and Bifulco, Roberto},
  booktitle={Proceedings of the 2025 Conference of the Nations of the Americas Chapter of the Association for Computational Linguistics: Human Language Technologies},
  pages={1543--1558},
  year={2025}
}

@book{anderson1976language,
  title={Language, Memory, and Thought},
  author={Anderson, J.R.},
  isbn={9780898591071},
  lccn={76021791},
  series={Experimental psychology series},
  year={1976},
  publisher={L. Erlbaum Associates}
}

@article{xu2025chain,
  title={Chain of draft: Thinking faster by writing less},
  author={Xu, Silei and Xie, Wenhao and Zhao, Lingxiao and He, Pengcheng},
  journal={arXiv preprint arXiv:2502.18600},
  year={2025}
}

@article{lee2025well,
  title={How well do llms compress their own chain-of-thought? a token complexity approach},
  author={Lee, Ayeong and Che, Ethan and Peng, Tianyi},
  journal={arXiv preprint arXiv:2503.01141},
  year={2025}
}

@article{chiang2024over,
  title={Over-reasoning and redundant calculation of large language models},
  author={Chiang, Cheng-Han and Lee, Hung-yi},
  journal={arXiv preprint arXiv:2401.11467},
  year={2024}
}

@article{nayab2024concise,
  title={Concise thoughts: Impact of output length on llm reasoning and cost},
  author={Nayab, Sania and Rossolini, Giulio and Simoni, Marco and Saracino, Andrea and Buttazzo, Giorgio and Manes, Nicolamaria and Giacomelli, Fabrizio},
  journal={arXiv preprint arXiv:2407.19825},
  year={2024}
}

@article{zhong2023improving,
  title={Improving task generalization via unified schema prompt},
  author={Zhong, Wanjun and Gao, Yifan and Ding, Ning and Liu, Zhiyuan and Zhou, Ming and Wang, Jiahai and Yin, Jian and Duan, Nan},
  journal={AI Open},
  volume={4},
  pages={120--129},
  year={2023},
  publisher={Elsevier}
}

@article{ma2025large,
  title={Large Language Models Meet Knowledge Graphs for Question Answering: Synthesis and Opportunities},
  author={Ma, Chuangtao and Chen, Yongrui and Wu, Tianxing and Khan, Arijit and Wang, Haofen},
  journal={arXiv preprint arXiv:2505.20099},
  year={2025}
}

@article{pan2024unifying,
  title={Unifying large language models and knowledge graphs: A roadmap},
  author={Pan, Shirui and Luo, Linhao and Wang, Yufei and Chen, Chen and Wang, Jiapu and Wu, Xindong},
  journal={IEEE Transactions on Knowledge and Data Engineering},
  volume={36},
  number={7},
  pages={3580--3599},
  year={2024},
  publisher={IEEE}
}

@article{kau2024combining,
  title={Combining knowledge graphs and large language models},
  author={Kau, Amanda and He, Xuzeng and Nambissan, Aishwarya and Astudillo, Aland and Yin, Hui and Aryani, Amir},
  journal={arXiv preprint arXiv:2407.06564},
  year={2024}
}

@article{lewis2020retrieval,
  title={Retrieval-augmented generation for knowledge-intensive nlp tasks},
  author={Lewis, Patrick and Perez, Ethan and Piktus, Aleksandra and Petroni, Fabio and Karpukhin, Vladimir and Goyal, Naman and K{\"u}ttler, Heinrich and Lewis, Mike and Yih, Wen-tau and Rockt{\"a}schel, Tim and others},
  journal={Advances in neural information processing systems},
  volume={33},
  pages={9459--9474},
  year={2020}
}
